\pdfoutput=1
\documentclass[11pt]{article}
\usepackage[]{EACL2023}

\usepackage{times}
\usepackage{latexsym}
\usepackage[T1]{fontenc}
\usepackage[utf8]{inputenc}
\usepackage{microtype}
\usepackage{url}
\usepackage{amsmath,graphicx}
\usepackage{amssymb,amsfonts}
\usepackage{textcomp,booktabs}
\usepackage{xcolor}
\usepackage{mathtools}
\usepackage{subfigure}
\usepackage{multirow}

\usepackage{amsmath,amsfonts,bm}

\def\eqref#1{equation~\ref{#1}}
\def\1{\bm{1}}

\def\vh{{\bm{h}}}

\def\vp{{\bm{p}}}

\def\vr{{\bm{r}}}
\def\vs{{\bm{s}}}

\def\vv{{\bm{v}}}

\def\mC{{\bm{C}}}

\def\mI{{\bm{I}}}

\def\mS{{\bm{S}}}

\def\mV{{\bm{V}}}

\DeclareMathAlphabet{\mathsfit}{\encodingdefault}{\sfdefault}{m}{sl}
\SetMathAlphabet{\mathsfit}{bold}{\encodingdefault}{\sfdefault}{bx}{n}

\def\gL{{\mathcal{L}}}

\def\gR{{\mathcal{R}}}

\def\gV{{\mathcal{V}}}

\def\sI{{\mathbb{I}}}

\newcommand{\R}{\mathbb{R}}

\DeclareMathOperator*{\cnn}{CNN}
\DeclareMathOperator*{\linear}{LinearLayer}
\DeclareMathOperator*{\mlp}{MLP}
\DeclareMathOperator*{\selfatten}{Self-Attention}
\DeclareMathOperator*{\generator}{Trans-Dec}
\DeclareMathOperator*{\lmhead}{LM-Head}
\DeclareMathOperator*{\simf}{sim}
\DeclareMathOperator*{\pool}{Pooling}
\DeclareMathOperator*{\mlptri}{MLP_{triplet}}
\DeclareMathOperator*{\rcider}{R_{CIDEr}}

\title{Style-Aware Contrastive Learning for Multi-Style Image Captioning}

\author{Yucheng Zhou, Guodong Long \\
         Australian AI Institute, School of Computer Science, FEIT, University of Technology Sydney \\
         {\tt yucheng.zhou-1@student.uts.edu.au, guodong.long@uts.edu.au}\\
         }

\begin{document}
\maketitle
\begin{abstract}
Existing multi-style image captioning methods show promising results in generating a caption with accurate visual content and desired linguistic style. However, existing methods overlook the relationship between linguistic style and visual content. To overcome this drawback, we propose style-aware contrastive learning for multi-style image captioning. First, we present a style-aware visual encoder with contrastive learning to mine potential visual content relevant to style. Moreover, we propose a style-aware triplet contrast objective to distinguish whether the image, style and caption matched. To provide positive and negative samples for contrastive learning, we present three retrieval schemes: object-based retrieval, RoI-based retrieval and triplet-based retrieval, and design a dynamic trade-off function to calculate retrieval scores. Experimental results demonstrate that our approach achieves state-of-the-art performance. In addition, we conduct an extensive analysis to verify the effectiveness of our method.
\end{abstract}

\section{Introduction}\label{sec:intro}
Stylized image captioning aims to generate a natural language description with stylized elements for a given image \cite{mathews2016senticap,gan2017stylenet}. With the advance of deep learning in human-computer interaction equipment, it has been integrated into many real-world applications like education robots \cite{Zhou22Sketch}, visual dialog \cite{Das17Visual}, and vision-language navigation \cite{Wang19Reinforced}. Therefore, it has attracted more attention from academia and industry and has become one of the essential areas in the natural language processing (NLP) and computer vision (CV) community. 

\begin{figure}[t]
  \centering
  \includegraphics[width=\linewidth]{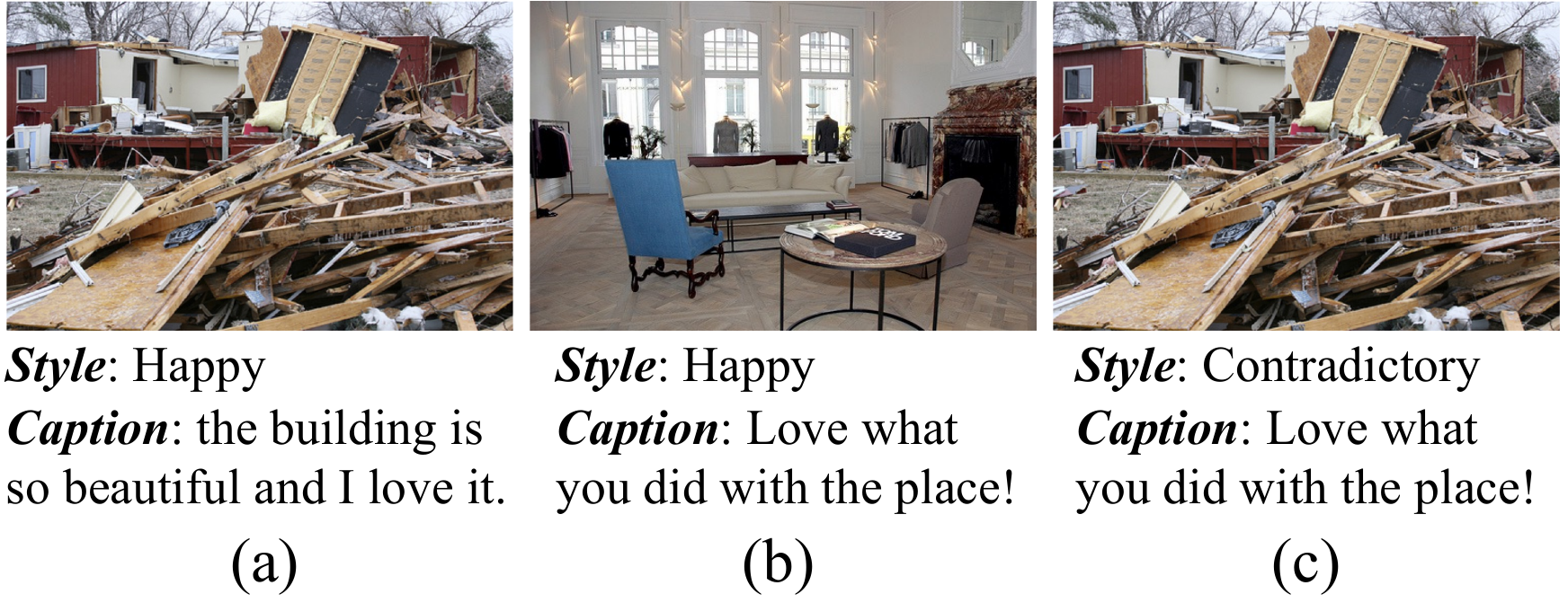}
  \caption{(a) the caption is generated by stylized captioning model \cite{shuster2019engaging}; (b) and (c) show the same caption can correspond to different styles.}
  \label{fig:intro}
\end{figure}

However, many methods propose translating images into captions of a single caption style \cite{mathews2016senticap,gan2017stylenet}. These methods need to train multiple models to handle multiple styles, which is very inefficient. Therefore, a rising demand for stylized image captioning is to learn an efficient model to handle multiple styles simultaneously. Recently, many efforts have been made in multi-style image captioning \cite{Guo_2019_CVPR,shuster2019engaging,Li213M}. 

Despite their success, existing methods suffer from a drawback: They focus on accurate visual content and desired linguistic style but overlook the relation between linguistic style and visual content. As shown in Figure \ref{fig:intro}(a), the generated caption contains accurate visual content and the desired style, but with misinformation that is easily detectable by humans, i.e., model generates ``beautiful'' in the caption to cater to the style of ``happy''. Indeed, linguistic style \cite{bell1984language} reflects personality, emotion and sentiment. In human behavior, these factors significantly influence the course of cognitive behavior \cite{simon_motivational_1967}. When people with different emotions see the same image are likely to describe different contents because they pay attention to different aspects. For example, people who are happy with animals may describe an image of a dog as a pretty dog with sparkling eyes and supple hair. However, one afraid of dogs may describe it as a scary dog with fierce teeth and sharp claws. The former focus on the dog's eyes and hair, while the latter on the dog's teeth and paws. Therefore, people with different emotions focus on different potential visual content. To generate human-like stylized captions, a stylized captioning model should learn to mine potential visual content relevant to different linguistic styles.

Due to the success of contrastive learning \cite{He20Momentum,Gao21SimCSE}, some works (e.g., UNITER \cite{Chen20UNITER}, ViLT \cite{Kim21ViLT}) employ contrastive learning objectives to encourage cross-modality alignment. In this work, we propose a style-aware visual encoder with contrastive learning to mine potential visual content relevant to styles. Specifically, style-aware visual encoder first mines potential visual features based on given image and style pairs. Then, we use a contrastive loss to pull potential visual features of the anchor and positive pair together while pushing those of anchor and negative pairs apart.

In addition, apart from requiring generated captions with accurate visual content and desired linguistic style, multi-style image captioning also needs to ensure that potential visual content and style are relevant. Moreover, since multi-style image captioning contains more fine-grained styles than single-stylized image captioning, it is difficult to directly distinguish the style by the caption. As shown in Figure \ref{fig:intro}(b) and (c), captions may be the same for two different styles. Therefore, multi-style image captioning is required to consider if matching among image, style and caption. Different from previous works \cite{Guo_2019_CVPR} that optimize generated captions only based on style, we propose a style-aware triplet contrast loss, which can learn the triplet matching among image, style, and caption by contrasting it with the positive triplet against negative ones.

Moreover, motivated by hard negatives sampling in retriever training \cite{Zhan21Optimizing}, we present three retrieval schemes to mine positive and negative examples for contrastive learning: object-based retrieval, RoI-based retrieval and triplet-based retrieval. These three schemes calculate scores according to object overlap rate, potential visual feature similarity, and triple feature similarity, respectively. Meanwhile, we design a dynamic trade-off function to calculate retrieval scores and analyze the impact of different retrieval schemes.

We conduct an extensive evaluation on three stylized image captioning datasets, i.e., SentiCap \cite{mathews2016senticap}, FlickrStyle10K \cite{gan2017stylenet} and PERSONALITY-CAPTIONS \cite{shuster2019engaging}. Our method significantly outperforms other strong competitors and achieves state-of-the-art performance. 

\section{Related Work} \label{sec:related_work}
\subsection{Image Captioning}
Image captioning, one of the essential tasks in multimodal research, aims to generate a description for a given image \cite{hodosh2013framing}. With the progress of deep learning, many end-to-end deep neural networks for image captioning are proposed \cite{vinyals2015show,anderson2018bottom,Yucheng21Triple}. Although these works have achieved excellent improvements, the caption generated by these models focuses on a single language domain, which means that their outputs can be in only one style and even dull and lack vitality sometimes.

\subsection{Stylized Image Captioning}
With the advance in image captioning techniques, researchers have attempted to generate an image caption with style. \citet{mathews2016senticap} propose a word-level regularization for captioner to model sentiment words. \citet{gan2017stylenet} employ transforming word embeddings matrices to control style factors for generated captions. To accurately describe visual content and reflect the desired linguistic style, some methods \cite{mathews2018semstyle,Zhao20MemCap} split a  stylized sentence into a style-related part that reflects the linguistic style and a content-related part that contains the visual content. \citet{Zhou22Sketch} employ prompt-based pre-training to build a stylized captioner without any paired sketch and story. These methods alleviate the reliance on paired training data for stylized captioner training. However, these methods need to train multiple models to handle multiple styles, which is inefficient. Therefore, learning an efficient model to handle multiple styles simultaneously raises more interest. Some efforts are made on multi-style image captioning, including adversarial learning network \cite{Guo_2019_CVPR} and multi-updown fusion model \cite{Li213M}. To generate more diverse outputs, \citet{shuster2019engaging} collect PERSONALITY-CAPTIONS, a large-scale multi-style image captioning dataset.

\subsection{Contrastive Learning}
Recently, contrastive learning has made exciting progress in representation learning \cite{He20Momentum,Gao21SimCSE,Zhou22ClarET}. \citet{He20Momentum} leverage contrastive learning to improve visual presentation learning in an unsupervised manner. \citet{Gao21SimCSE} propose a simple unsupervised contrastive learning method to perform on par with previously supervised counterparts. \citet{Yang22Vision-Language} propose triple contrastive learning for vision-language pre-training by leveraging both cross-modal and intra-modal self-supervision, providing complimentary benefits in representation learning. These works show the powerful ability of contrastive learning to improve representation learning.

\section{Method}
\begin{figure}[t]
\centering
\includegraphics[width=\linewidth]{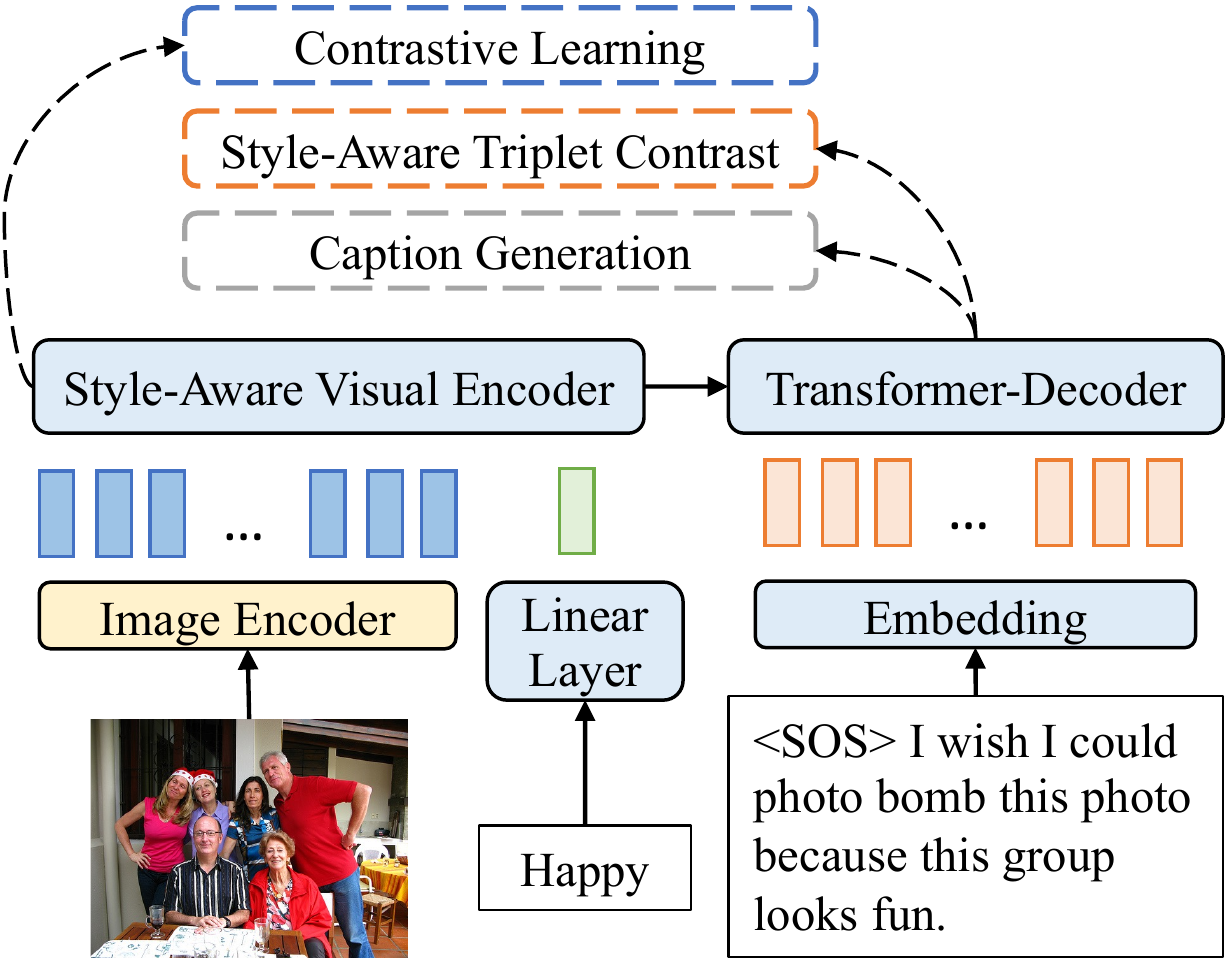}
\caption{An overview of our proposed SACO model with three objectives. Yellow rounded rectangles denote fixed model. Blue rounded rectangles denote parameters that will be optimized.}
\label{fig:model}
\end{figure}

This section will elaborate on our \textbf{S}tyle-\textbf{A}ware \textbf{CO}ntrastive learning (\textbf{SACO}) method for multi-style image captioning, followed by our proposed novel retrieval schemes. The details of our approach are shown in Figure~\ref{fig:model}.  Lastly, details about training and fine-tuning are elaborated.

\subsection{Image and Style Encoding}
Firstly, we pass an image $\mI$ into a pre-trained convolutional neural network (CNN) to extract its visual feature $\mV$ and convert the visual feature into a sequence according to the row-first direction:
\begin{align}
\notag \mV &= \cnn(\mI) \\
  &\text{where}~ \mV =  [\vv_0, \vv_1, ..., \vv_m], \vv_i \in \R^{|d'|} \label{equ:enc}
\end{align}

In addition, the style $\mS$ is represented as a one-hot vector and encoded to a style feature $\vs$ by a linear layer:
\begin{align}
\vs = \linear(\mS), \vs \in \R^{|d|}
\end{align}

Since there is a different dimension between visual feature $\mV$ and style feature $\vs$, a multi-layer perceptron (MLP) built upon $\vv_i$ is presented to reduce its dimension to equal the dimension of $\vs$:
\begin{align}
\vv_i := \mlp(\vv_i), ~\vv_i \in \R^{|d|}, ~\forall i = 1,\dots, m
\end{align}

\subsection{Style-Aware Visual Encoder}\label{sec:svc}
Originating from the observation that different styles will focus on different image regions, we propose a style-aware visual encoder with contrastive learning to mine potential visual content relevant to styles. Since self-attention \cite{VaswaniSPUJGKP17} shows a strong capability to relate different positions of a sequence for feature computation, we leverage self-attention as backbone of the style-aware visual encoder. Particularly, we concatenate the visual feature $\mV$ and style feature $\vs$, and apply the self-attention layers to them to derive the style-aware visual features $\mV^s$ and vision-aware style feature $\vs^v$, i.e.,
\begin{align}
[\mV^s;\vs^v] = \selfatten([\mV;\vs]) \label{equ:selfatten}
\end{align}
where $[\cdot;\cdot]$ denotes the operation of concatenation. Intuitively, the success of Eq.~(\ref{equ:selfatten}) requires accurately capturing the visual features relevant to styles, but a major problem arises without artifact labels: \textit{Learning Difficulty}. Motivated by some works \cite{Chen20UNITER,Kim21ViLT} implement cross-modality alignment via contrastive learning, we leverage contrastive learning for style-aware visual encoder, which drives its learning to capture potential visual content correlated to given styles without any labeled data. Specifically, we use contrastive learning to learn the representation of potential visual content by contrasting it with the positive example $(\hat \mV^s, \hat \vs^v)$ against those of negative ones $(\bar \mV^s_i, \bar \vs^v_i)$. Particularly, we first derive features $(\hat \mV^s, \hat \vs^v)$ and $(\bar \mV^s_i, \bar \vs^v_i)$ independently via Eq. (\ref{equ:enc}-\ref{equ:selfatten}). Then, we mine more accurate $(\mV^s, \vs^v)$, style-aware visual features and vision-aware style features, by contrasting it with $(\hat \mV^s, \hat \vs^v)$ against $(\bar \mV^s_i, \bar \vs^v_i)$, i.e.,
\begin{align} 
\notag\gL^{(svc)} \!\!=\!\! -\log &\frac{e^{\simf(r,\hat r) / \tau}}{e^{\simf(r,\hat r) / \tau} \!\!+\!\! \sum_{i=1}^{M}e^{\simf(r, \bar r_i) / \tau}} \\
\notag -\log &\frac{e^{\simf(s^v,\hat s^v) / \tau}}{e^{\simf(s^v,\hat s^v) / \tau} \!\!+\!\! \sum_{i=1}^{M}e^{\simf(s^v, \bar s^v_i) / \tau}} \\
\notag \text{where}~&r=\pool(\mV^s)  \\     
\notag &\hat r=\pool(\hat \mV^s)  \\     
&\bar r=\pool(\bar \mV^s_i)
\label{equ:svc}
\end{align}
where $\simf(\vr, \vr')$ is the dot product operation between $\ell_{2}$ normalized $\vr$ and $\vr'$ (i.e. cosine similarity $\frac{\vr^{\top} \vr'}{\left\|\vr\right\| \cdot\left\|\vr'\right\|}$) and $\tau$ is a temperature hyperparameter to control the pull and push force; $M$ is number of negative samples. As a result, since the style-aware visual features and vision-aware style features also offer a straightforward pathway to transmit style-aware visual information to style-aware visual encoder, it mitigates the \textit{learning difficulty} problem.

\subsection{Transformer-Decoder Based Generator}
Due to the success of the pre-trained language model, there are some works (e.g., GPT \cite{radford2018improving}, GPT-2 \cite{radford2019language}) pre-train the Transformer decoder~\cite{VaswaniSPUJGKP17} on large-scale corpora. Recently, a fundamental paradigm of text generation tasks is to fine-tune the pre-trained model on the target data, and it can achieve exciting performance. In this work, we leverage a trained transformer decoder to initialize our generator. The generator input is adjusted to the triples $(\mV^s, \vs^v, \bar{\mC})$, where $\bar{\mC}$ refers to the segment of stylized image caption. 
The purpose of generator is to predict a probability distribution of the next word of the segment $\bar{\mC}$ based on the given triple, i.e.,
\begin{align}
\label{equ:h}\notag \vh_{i} &= \generator(\mV^s, \vs^v, \bar{\mC}) \in \R^{|d|} \\
    &\text{where}~\bar{\mC} = [c_1, \dots, c_{i-1}] \\ 
\vp_{i} &= \lmhead(\vh_{i}) \in \R^{|\gV|}
\end{align}
where $\vh_i$ refers to the hidden representation in $i$-th step; $\gV$ denotes token vocabulary and $\vp_{i}$ is a probability distribution over $\gV$. Lastly, the caption generation objective is defined as a maximum likelihood estimation and written as:
\begin{align}
    \gL^{(cap)} = - \dfrac{1}{|N|}\sum\nolimits_{i=1}^{N}\log \vp_{i}(c_i), \label{equ:loss_cap}
\end{align}
where $\vp_{i}(c_i)$ denotes fetching the probability of the $i$-th step gold token $c_i\in\mC$ from $\vp_{i}$. $\mC$ refers to the gold caption and $N$ is its length.

\subsection{Style-Aware Triplet Contrast}\label{sec:stc}
Different from stylized image captioning for a single style, multi-style image captioning contains more fine-grained styles, and it is difficult to distinguish the style by the caption directly. So multi-style image captioning is required to consider if matching among image, style and caption. Recently, contrastive learning has shown its power capability in alignment between positive pairs and dispersion between negative ones. Inspired by advances in contrastive learning, we propose a style-aware triplet contrast loss to learn the triplet matching among image, style, and caption by contrasting it with the positive triplet against negative ones. Given an image $\mI$, style $\mS$ and caption $\mC$, we first obtain the representation for this triplet. Since visual and style features have fed into the decoder, the triplet representation can be extracted by the hidden representation of the decoder, i.e., 
\begin{align}
  \vh = - \dfrac{1}{|N|}\sum\nolimits_{i=1}^{N} \mlptri(\vh_i) 
\end{align}
where $N$ is the same as that in Eq.\ref{equ:loss_cap} and denotes the length of the caption; $\mlptri$ refers to a Multilayer Perceptron; and $\vh_i$ can be derived from Eq.\ref{equ:h}.
Then, we enhance the triplet representation $\vh$ by contrasting it with a positive triplet $\hat \vh$ against negative ones $\bar \vh$, i.e.
\begin{align} 
  \gL^{(stc)} \!\!=\!\! -\log &\frac{e^{\simf(\vh,\hat \vh) / \tau}}{e^{\simf(\vh,\hat \vh) / \tau} \!\!+\!\! \sum_{i=1}^{M}e^{\simf(\vh, \bar \vh) / \tau}} \label{equ:stc}
\end{align}
where $\simf(\cdot, \cdot)$ is the dot product operation as same as that in Eq.\ref{equ:svc}.

\subsection{Retrieval Schemes}
In our proposed contrastive learning objective, positive and negative samples are important elements. Since caption datasets are not designed with positive and negative samples, we propose three heuristics to derive positive and negative samples for a triplet of image $\mI$, style $\mS$ and caption $\mC$:

\paragraph{Object-based Retrieval. }
We first leverage a well-trained object detection model \cite{Ren15Faster} to obtain object classes in images. Then, we retrieve image $\hat \mI$ from image set $\sI$ according to object overlap with $\mI$, and derive a probability for sampled examples:
\begin{align}
  P_{obj} = \dfrac{N_{overlap}}{N_{\mI}}
\end{align}
where $N_{\mI}$ denotes number of objects in the image $\mI$, and $N_{overlap}$ refers to the number of overlapped objects both in $\mI$ and $\hat \mI$.

\paragraph{RoI-based Retrieval. }
In this retrieval scheme, the region of interest (RoI) refers to potential visual content relevant to style. We retrieve image $\hat \mI$ based on the similarity between its representation of potential visual content and that of image $\mI$:
\begin{align}
  P_{roi} = \simf(\mV, \hat \mV)
\end{align}
where $\mV$ and $\hat \mV$ are the style-aware visual features of $\mI$ and $\hat \mI$, and details of their calculation can refer to Eq.\ref{equ:svc}.

\paragraph{Triplet-based Retrieval. }
Since triplet matching is essential for multi-style image captioning, we retrieve image $\hat \mI$ according to the similarity between triplets:
\begin{align}
  P_{tri} = \simf(\vh, \hat \vh)
\end{align}
where $\vh$ and $\hat \vh$ are the triplet representation for the triplet $(\mI, \mS, \mC)$ and $(\hat \mI, \hat \mS, \hat \mC)$, and details of their calculation can refer to Eq.\ref{equ:stc}.

\paragraph{Dynamic Trade-off Function. } \label{sec:sample}
We combine the above three novel retrieval schemes to rank samples, and score of each sample can be defined as:
\begin{align}
  P \!=\! \theta^{\mu}P_{obj} \!+\! (1 \!-\! \theta^{\mu})(\phi P_{roi} \!+\! (1 \!-\! \phi) P_{tri}) \label{equ:p_cal}
\end{align}
where $P$ denotes the score of samples and $\phi$ is a trade-off parameter; $\theta$ denotes a decay factor and $\mu$ is the current training epoch. During training, we randomly select a sample among the top-10 samples as positive one, and the negative samples are randomly sampled based on $P < max(0.1, P_{max} - \omega^{\mu})$.

\subsection{Training and Fine-tuning}
For model training, we adopt the same 2-stage training scheme (training and fine-tuning) as in \cite{anderson2018bottom}. In the training stage, we optimize the model according to the three objectives proposed above, and the loss function of our model can be integrated into the following:
\begin{align}
  \gL = \gL^{(cap)} + \alpha \gL^{(svc)} + \beta \gL^{(stc)} \label{equ:loss_total}
\end{align}
where $\alpha$ and $\beta$ are trade-off parameters.

In the fine-tuning stage, we employ the CIDEr score to optimize our model as same as \cite{rennie2017self}, i.e., returning a reward for generated caption $\hat \mC$.
\begin{align}
  \gR^{(CIDEr)} = \rcider(\hat \mC) - b
\end{align}
where $b$ is a baseline, i.e., the reward $\rcider(\mC^*)$ for the generated caption $\mC^*$ with greedy search.

\section{Experiments}
\subsection{Dataset and Evaluation Metrics}
We evaluate our proposed approach on three datasets, PERSONALITY-CAPTIONS \cite{shuster2019engaging}, SentiCap \cite{mathews2016senticap} and FlickrStyle10K \cite{gan2017stylenet}.

\paragraph{PERSONALITY-CAPTIONS. } \citet{shuster2019engaging} collect a large-scale multi-style image captioning dataset, PERSONALITY-CAPTIONS, which includes 201,858 images, 215 personality traits, and 241,858 stylized captions. We divide the dataset following \cite{shuster2019engaging}, and the size of the training set, validation set, and test set are 186,858, 5,000, and 10,000, respectively. In the test set, each image contains five reference captions. Following \citet{shuster2019engaging}, we use the same metrics to report our results, and the evaluation is based on the coco-caption code$\footnote{https://github.com/tylin/coco-caption}$. The evaluation metrics include: BLEU~\cite{papineni2002bleu}, ROUGE-L~\cite{lin2004rouge}, CIDEr~\cite{vedantam2015cider}, SPICE~\cite{anderson2016spice}.

\paragraph{SentiCap \& FlickrStyle10K. } SentiCap \cite{mathews2016senticap} and FlickrStyle10K \cite{gan2017stylenet} are two publicly stylized image caption datasets. According to \cite{Li21Similar}, we process SentiCap and FlickrStyle10K datasets, and use samples from MSCOCO \cite{Lin14Microsoft} and Flickr30K \cite{hodosh2013framing} as large-scale paired factual data. Following \citet{Li21Similar}, we use BLEU, METEOR \cite{banerjee2005meteor}, CIDEr, style classification accuracy (cls.) and the average perplexity (ppl.) as evaluation metrics. Style classification accuracy is measured by a well-trained BERT, which achieves accuracies of 95.9\%, 98.0\%, 98.1\%, and 99.5\% on the test sets of humorous, romantic, negative, and positive styles, respectively. The average perplexity is measured by a well-trained trigram-based statistical language model using SRILM toolkit. A lower score denotes that the generated caption is more fluent and reflects the desired linguistic style better.

\subsection{Implementation Details}
Our approach adopts the same 2-stage training scheme (training and fine-tuning).
For training stage, input images are resized to the size of $256 \times 256$, and then we randomly crop the image size as $224 \times 224$ as model input. 
To ensure a fair comparison, we use the ResNeXt-IG-3.5B~\cite{Mahajan_2018_ECCV} as the pre-trained image encoder, as same as \cite{shuster2019engaging}, and the size of output features is $7\times7\times2048$. 
Then, the visual features are reshaped as $49\times2048$ according to the row-first direction. 
The style-aware visual encoder is constructed with three self-attention layers. 
We use a distilled version of pre-trained GPT-2 \cite{sanh2019distilbert} as our transformer-decoder based generator, as same as \cite{Nguyen20Structural}. 
The layers and attention heads of the decoder are 6 and 8. 
The dimension of embedding vectors and hidden states in the decoder are 768 and 1024. 
Special tokens for the beginning and end of sentences are <SOS> and <EOS>. 
In addition, the size of the style linear layer are $215\times768$ and $5\times768$ for PERSONALITY-CAPTIONS dataset and SentiCap and FlickrStyle10K datasets. 
For model training, we utilize the Adam optimizer~\cite{KingmaB14} with learning rate of 1e-4. 
The batch size, warm-up proportion, weight decay, maximum training epoch and temperature hyperparameter $\tau$ are 128, 0.1, 0.01, 10 and 0.08. 
The trade-off parameter $\phi$ and decay factor $\theta$ in Eq.\ref{equ:p_cal} are 0.5 and 0.9. $\omega$ for negatives sampling is 0.8. Trade-off parameters $\alpha$ and $\beta$ in Eq.\ref{equ:loss_total} are 0.5 and 0.7. 
For fine-tuning stage, the maximum training epoch and learning rate are 3 and 1e-5. 
Other experimental details are the same as that of training stage. 
For testing, we use beam search with beam size of 3 to generate captions with maximum sentence length of 30. Our model is trained on one V100 GPU.

\subsection{Baselines}
We compare our model with the following state-of-the-art baselines: 
(1) {\bf ShowTell} propose by \cite{shuster2019engaging} and is a variant of \cite{vinyals2015show} that concatenates the style features with the input word vectors at each decoding step.
(2) {\bf ShowAttTell}, proposed by \cite{xu2015show}, aims to enhance the correlation between the text and image by an attention mechanism, and the used model is a variant proposed by \cite{shuster2019engaging}.
(3){\bf UpDown} with a decoder of two LSTMs can adapt to generate attention weights and use it to generate captions~\cite{anderson2018bottom}, and the used model is a variant proposed by \cite{shuster2019engaging}.
(4) {\bf GPT} with an image encoder is fine-tuned on the captioning dataset \cite{radford2019language}. 
(5) {\bf GPT-Speaker} \cite{Nguyen20Structural} employs the language model GPT2 as a language prior for both the speaker and listener in the multi-agent communication framework.
(6) {\bf 3M} \cite{Li213M} is a multi-style image captioner that is a multi-UpDown encoder-decoder model integrated with multi-modal features.
  
\subsection{Main Results}
\begin{table}[t] \small
  \centering
  \begin{tabular}{lccccc}
  \toprule
  \textbf{Method} & \textbf{B@1} & \textbf{B@4} & \textbf{R} & \textbf{C} & \textbf{S} \\
  \midrule
  ShowTell        & 38.4         & 7.3          & 24.3       & 9.6        & 1.6        \\
  ShowAttTell     & 43.3         & 7.1          & 27.0       & 12.6       & 3.6        \\
  UpDown          & 44.4         & 8.0          & 27.4       & 16.5       & 5.2        \\
  GPT             & 49.2         & 9.1          & 29.0       & 19.0       & 6.3        \\
  GPT-Speaker     & 52.1         & 8.4          & 30.2       & 19.9       & 7.3        \\
  GPT-Speaker*     & 52.3         & 8.2          & 30.1       & 20.0       & 7.4        \\
  3M              & 43.0         & 8.0          & 27.6       & 18.6       & 4.8        \\
  SACO (Ours)     & \bf{54.8}         & \bf{9.7}          & \bf{32.6}       & \bf{21.0}       & \bf{8.1}   
   \\
  \bottomrule
  \end{tabular}
  \caption{Comparison results on PERSONALITY-CAPTIONS test set. B@1, B@4, R, C and S denote BLEU@1, BLEU@4, ROUGE-L, CIDEr and SPICE, respectively. $*$ refers to the baseline of our reproduction.}
  \label{tab:exp}
\end{table}

\begin{table}[t]\small
\centering
\setlength\tabcolsep{3.2pt}
\begin{tabular}{llcccccc}
\toprule
\textbf{Style}         & \textbf{Method} & \textbf{B@1}  & \textbf{B@3}  & \textbf{M}    & \textbf{C}    & \textbf{cls.}  & \textbf{ppl.} \\
\midrule
\multirow{4}{*}{Humor} & SF-LSTM         & 27.4          & 8.5           & 11.0          & 39.5          & -              & -             \\
                       & MSCap           & 16.3          & 1.9           & 5.3           & 15.2          & 91.3           & 22.7          \\
                       & SAN             & 29.5          & 9.9           & 12.5          & 47.2          & 99.4           & 13.7          \\
                       & SACO            & \textbf{35.2} & \textbf{10.3} & \textbf{13.4} & \textbf{51.2} & \textbf{99.7}  & \textbf{12.9} \\
\midrule
\multirow{4}{*}{Roman} & SF-LSTM         & 27.8          & 8.2           & 11.2          & 37.5          & -              & -             \\
                       & MSCap           & 17.0          & 2.0           & 5.4           & 10.1          & 88.7           & 20.4          \\
                       & SAN             & 30.9          & 10.9          & 13.0          & 53.3          & 99.6           & 13.1          \\
                       & SACO            & \textbf{35.8} & \textbf{13.5} & \textbf{14.1} & \textbf{55.8} & \textbf{99.9}  & \textbf{12.4} \\
\midrule
\multirow{4}{*}{Pos}   & SF-LSTM         & 50.5          & 19.1          & 16.6          & 60.0          & -              & -             \\
                       & MSCap           & 46.9          & 16.2          & 16.8          & 55.3          & 92.5           & 19.6          \\
                       & SAN             & 53.0          & 23.4          & 18.1          & 72.0          & \textbf{100.0}          & 11.7          \\
                       & SACO            & \textbf{56.3} & \textbf{24.7} & \textbf{19.5} & \textbf{72.2} & 99.8 & \textbf{11.5} \\
\midrule
\multirow{4}{*}{Neg}   & SF-LSTM         & 50.3          & 20.1          & 16.2          & 59.7          & -              & -             \\
                       & MSCap           & 45.5          & 15.4          & 16.2          & 51.6          & 93.4           & 19.2          \\
                       & SAN             & 51.2          & 20.5          & 17.6          & 67.0          & 100.0 & 14.8          \\
                       & SACO            & \textbf{53.2} & \textbf{23.6} & \textbf{19.1} & \textbf{70.5} & \textbf{100.0}           & \textbf{13.3}   \\
\bottomrule
\end{tabular}
\caption{Comparison results on FlickrStyle10K and SentiCap test set. M denotes METEOR.}
\label{tab:sec}
\end{table}
Comparison results on PERSONALITY-CAPTIONS test set are shown in Table \ref{tab:exp}. From the table, we can make three observations. First, we can observe that our methods achieve state-of-the-art performance on the PERSONALITY-CAPTIONS dataset. Second, models based on GPT reach a better performance than LSTM-based models, which shows the powerful generation capability of transformer-based decoder in stylized caption generation. Lastly, our model significantly outperforms the GPT-2 model by a large margin (i.e., improving 2.0 in CIDEr). That is, style-aware contrastive learning can improve the model by strengthening different style understanding.

In addition, we also show comparison results on FlickrStyle10K and SentiCap test set in Table \ref{tab:sec}. Competitors include SF-LSTM \cite{Chen18Factual}, MSCap \cite{Guo_2019_CVPR} and SAN \cite{Li21Similar}. Results show that transformer-based methods (i.e., SAN and SACO) outperform LSTM-based methods (i.e., SF-LSTM and MSCap). Moreover, compared with strong competitors, our approach achieves state-of-the-art performance on FlickrStyle10K and SentiCap.

\subsection{Ablation Study}
\begin{table}[t] \small
  \centering
  \begin{tabular}{lccccc}
  \toprule
  \textbf{Method}     & \textbf{B@1} & \textbf{B@4} & \textbf{R} & \textbf{C} & \textbf{S} \\
  \midrule
  SACO            & \bf{54.8}         & \bf{9.7}          & \bf{32.6}       & \bf{21.0}       & \bf{8.1}           \\
  $\Diamond$ w/o CIDEr           & 53.0         & 9.6          & 31.9       & 19.1       & 7.2        \\
  $\Diamond$ w/o STC             & 52.9         & 9.0          & 31.2       & 20.0       & 7.5        \\
  $\Diamond$ w/o SVC             & 52.4         & 8.7          & 30.8       & 19.8       & 7.3        \\
  $\Diamond$ w/o SVC, STC        & 48.9         & 9.0          & 29.1       & 18.7       & 6.0        \\
  $\Diamond$ $\gL^{(cap)}$ Only  & 47.3         & 8.7          & 29.2       & 16.0       & 5.4        \\
  \bottomrule
  \end{tabular}
  \caption{\small Ablation study. 
  ``\textit{w/o CIDEr}'' denotes removing fine-tuning stage for model training; 
  ``\textit{w/o STC}'' denotes removing the style-aware triplet contrast loss; 
  ``\textit{w/o SVC}'' denotes removing the style-aware visual contrast encoder; 
  ``\textit{w/o SVC,STC}'' denotes removing both contrastive learning methods in our model; 
  and ``\textit{$\gL^{(cap)}$ Only}'' denotes our model only train with the caption generation objective and without fine-tuning.}
  \label{tab:abl}
\end{table}
To demonstrate the effectiveness of our method, we conduct an ablation study and results are shown in Table \ref{tab:abl}. We first investigate the impact of the style-aware triplet contrast objective by removing it and find that model performance is decreased. Next, we investigate our method without style-aware visual contrast encoder and observe that the performance drops, which is worse than that without SVC. Moreover, we remove both contrastive learning objectives, and results show the performance is degraded again. The observations above demonstrate the effectiveness of style-aware contrastive learning.

\subsection{Analysis}

\subsubsection{Impact of Contrastive Learning}
\paragraph{Impact of Style-Aware Visual Contrast Encoder. }
\begin{table}[t] \small
  \centering
  \setlength\tabcolsep{4pt}
  \begin{tabular}{lccccc}
  \toprule
  \textbf{Method}      & \textbf{B@1} & \textbf{B@4} & \textbf{R} & \textbf{C} & \textbf{S} \\
  \midrule
  SACO            & \bf{54.8}         & \bf{9.7}          & \bf{32.6}       & \bf{21.0}       & \bf{8.1}        \\
  SACO (Dec w/o style) & 53.8         & 9.4          & 31.9       & 20.5       & 7.8        \\
  $\Diamond$ w/o SVC              & 52.9         & 9.0          & 31.2       & 20.0       & 7.5        \\
  \bottomrule
  \end{tabular}
  \caption{Impact of style-aware visual contrast encoder.}
  \label{tab:pvc}
\end{table}
To further investigate the impact of the style-aware visual contrast encoder, we remove the input style for our decoder, i.e., the decoder input only consists of $\mV^s, \bar{\mC}$ in Eq.\ref{equ:h}. Results are shown in Table \ref{tab:pvc}. Results show that our decoder without input style outperforms the model without SVC, which demonstrates that style-aware visual contrast encoder can capture potential visual content relevant to the given style.

\paragraph{Impact of Style-Aware Triplet Contrast. }
\begin{table}[t] \small
  \centering
  \setlength\tabcolsep{4pt}
  \begin{tabular}{lccccc}
  \toprule
  \textbf{Method}      & \textbf{B@1} & \textbf{B@4} & \textbf{R} & \textbf{C} & \textbf{S} \\
  \midrule
  SACO            & \bf{54.8}         & \bf{9.7}          & \bf{32.6}       & \bf{21.0}       & \bf{8.1}        \\
  SACO (STC-pointwise) & 53.2         & 9.0          & 31.4       & 20.1        & 7.3        \\
  SACO (comp.)          & 53.0         & 9.1          & 31.2       & 20.2        & 7.4        \\
  $\Diamond$ w/o STC             & 52.4         & 8.7          & 30.8       & 19.8       & 7.3        \\
  \bottomrule
  \end{tabular}
  \caption{Impact of style-aware triplet contrast.}
  \label{tab:ptc}
\end{table}
To investigate the impact of the style-aware triplet contrast objective, we replace the objective with two binary classification based objectives: point-wise classification and the comp. objective in \cite{Nguyen20Structural}. As shown in Table \ref{tab:ptc}, results show that binary classification based objectives underperform our contrastive objective, which demonstrates that contrasting positive and negative triplets is helpful for the model to understand triplet matching.

\subsubsection{Retrieval Schemes Ablation}
\begin{figure}[t]
\centering
\includegraphics[width=0.48\linewidth]{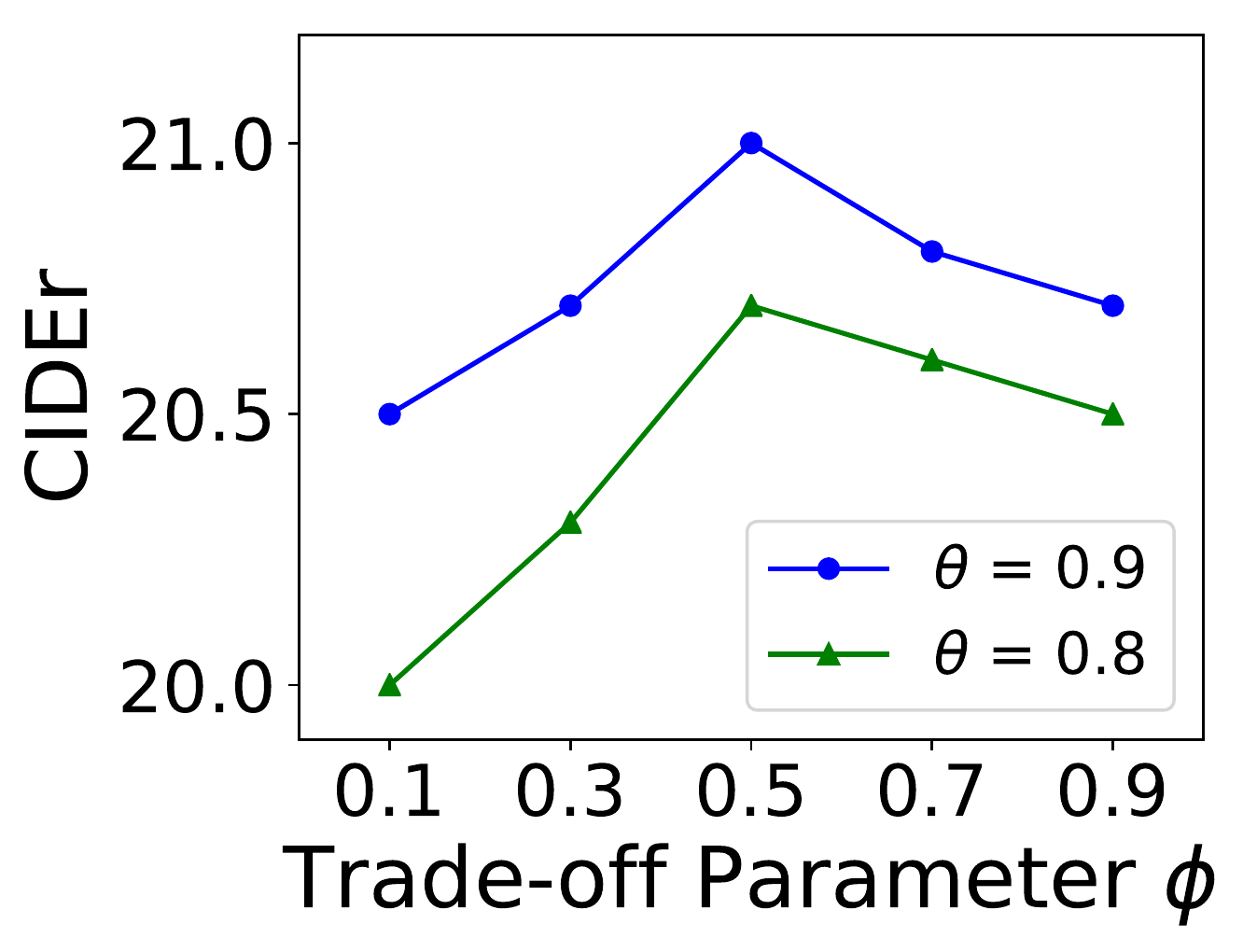}  
\includegraphics[width=0.48\linewidth]{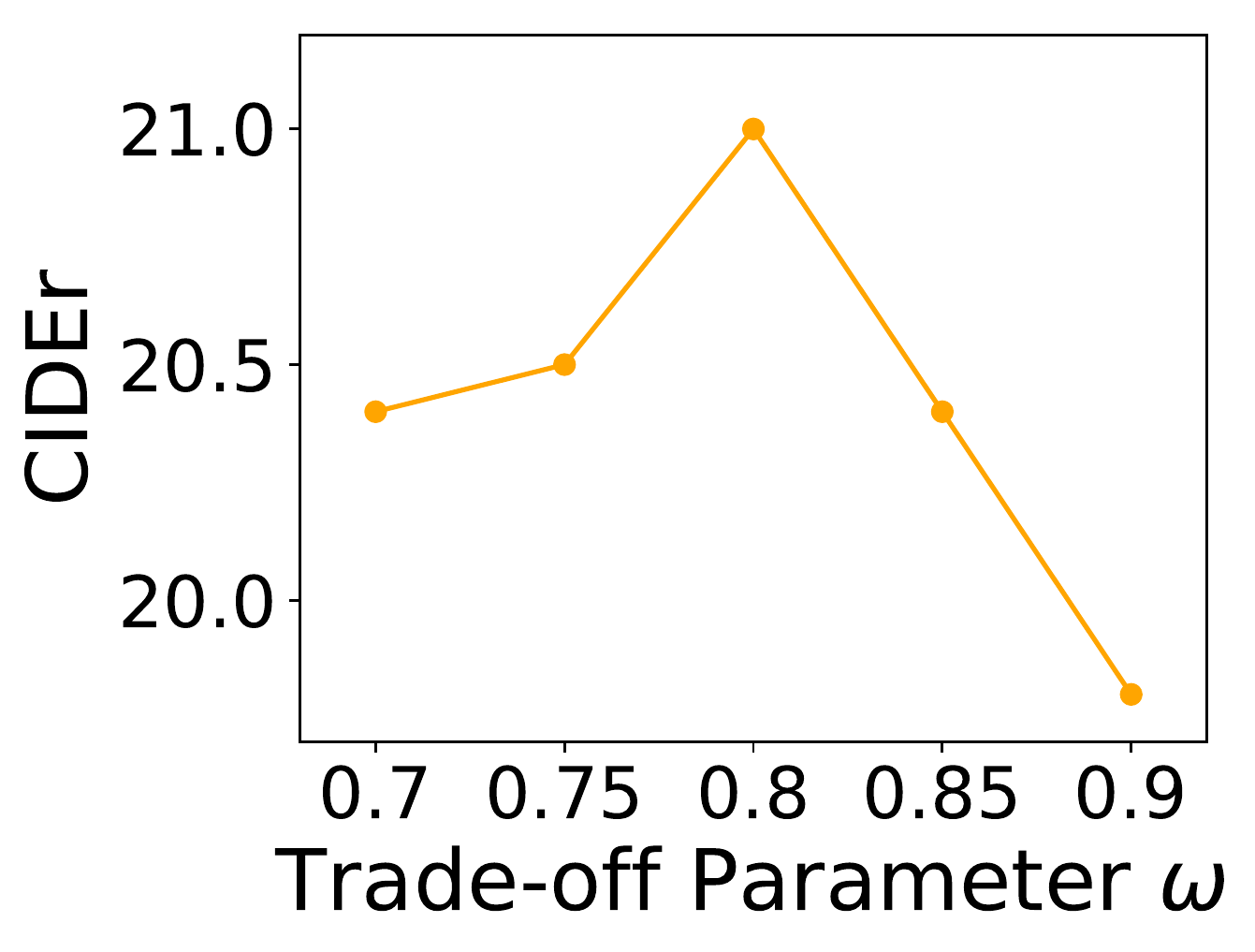}  
\caption{Retrieval schemes ablation.}
\label{fig:cider}
\end{figure}
Due to contrastive learning depending on positive and negative sampling, we conduct an ablative analysis on retrieval schemes. Results are shown in Figure \ref{fig:cider}. From left figure, we can make two observations: 
(1) The decay factor $\theta$ set of 0.9 can perform better than that of 0.8. It demonstrates that object-based retrieval is essential in the early training phase.
(2) The trade-off parameter $\phi$ set of 0.5 can achieve the best performance, which shows the vital role of both style-aware contrastive objectives in positive and negative sampling.
From right figure, based on the sampling function in \S \ref{sec:sample}, as $\omega$ becomes greater, negative sampling range during training will become greater. We can see that a greater negative sampling range is beneficial for contrasting learning. The reason is that the larger sampling range allows model to access harder negatives. In addition, we find that the performance gradually drops when $\omega$ is greater than 0.8, which shows too hard negatives hinder model learning.

\subsubsection{Qualitative Comparison}
\begin{figure}[t]
  \centering
  \includegraphics[width=\linewidth]{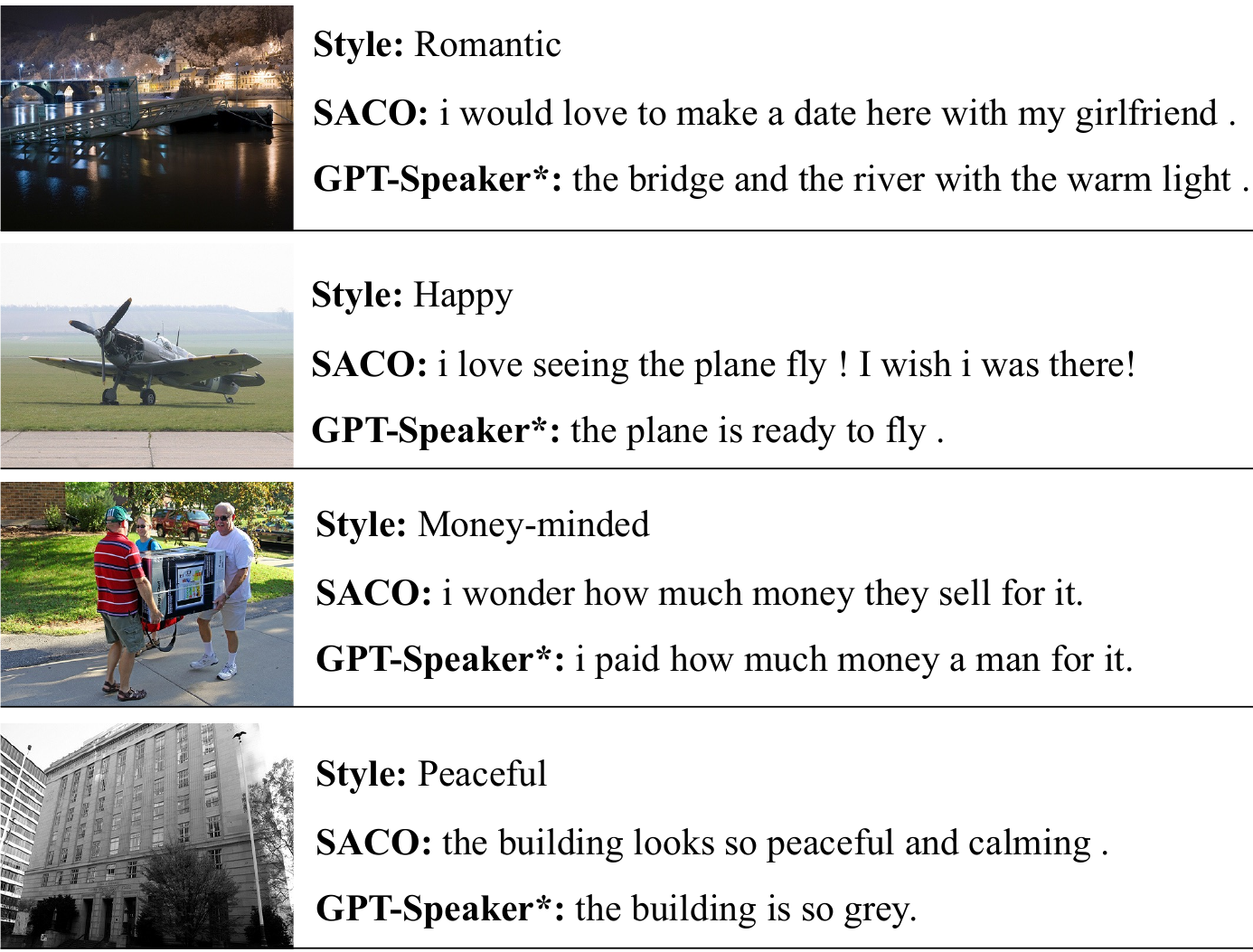}  
\caption{Random sampling examples generated by SACO and GPT-Speaker*.}
\label{fig:per}
\end{figure}
To extensively evaluate our model, we conduct a qualitative comparison of our model and GPT-Speaker, and some random sampling examples are shown in Figure~\ref{fig:per}. For example, in the first line, we can observe that GPT-Speaker can capture objects in the image and describe them in a caption, but it does not entail style. Therefore, the caption generated by our model is shown to more natural and with desired style. For instance, in the second and last lines, we can find that the caption generated by GPT-Speaker is more like a factual description. In contrast, the caption generated by our model is more expressive of style. 

\subsubsection{Interpretable Visualization Analysis}
\begin{figure}[t]
  \centering
  \includegraphics[width=\linewidth]{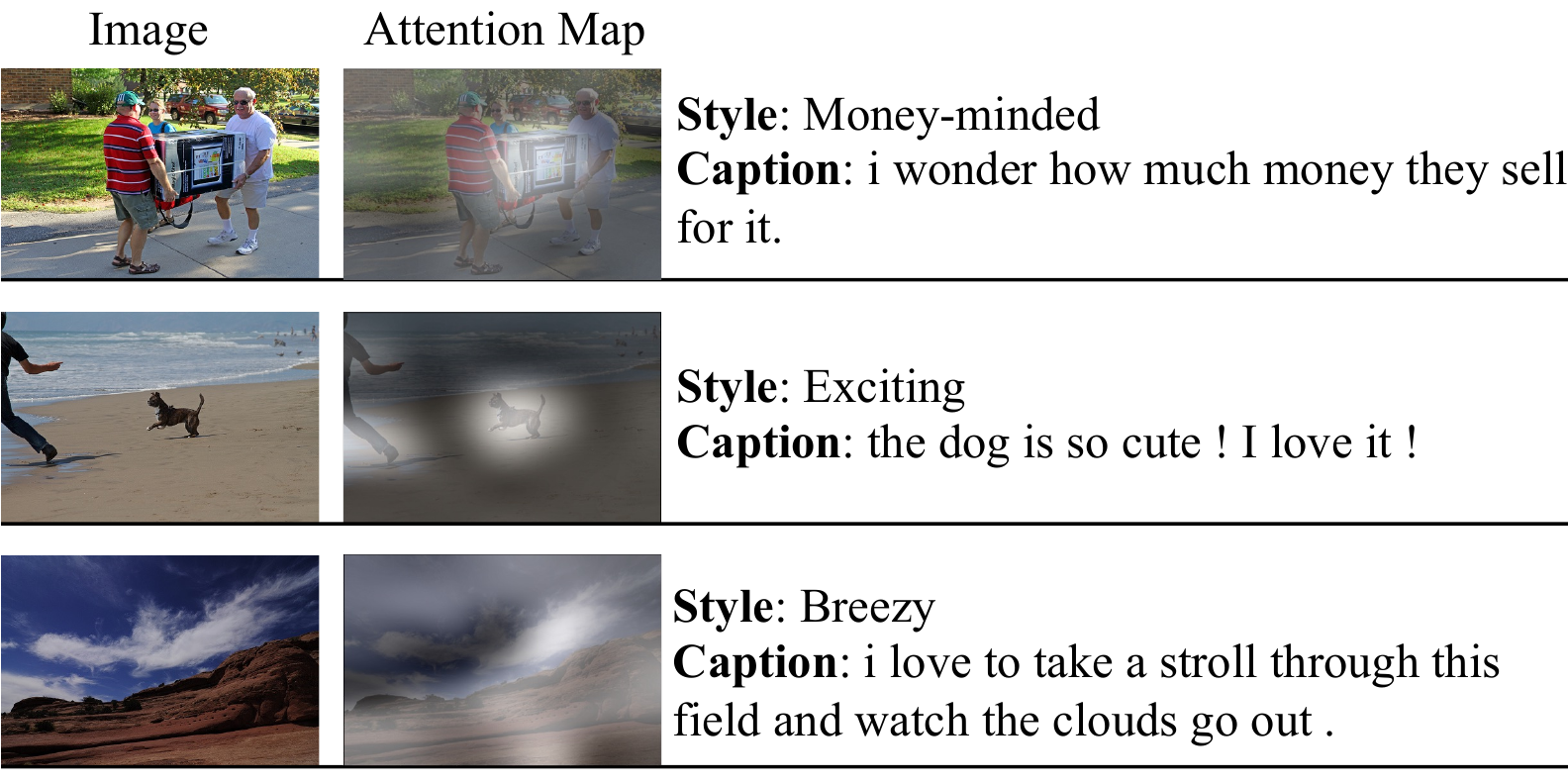}  
\caption{Interpretable visualization analysis of random sampling examples. The brighter image areas mean more relevant to style.}
\label{fig:vis}
\end{figure}
To investigate the effectiveness of style-aware visual contrast encoder, we conduct an interpretable visualization analysis, as shown in Figure \ref{fig:vis}. In the attention map of self-attention layers, the brighter image areas mean greater attention weights, i.e., these areas are more relevant to the given style. As shown in the first example,  under the style "money-minded", the object is more paid attention to than the human, which is very intuitive. Next, in the second example, under the style  "exciting", "a dog of running" are paid more attention and described with "love" and "cute", which are reasonable. Therefore, the results show that the style-aware visual contrast encoder can effectively capture potential visual content related to style.

\subsubsection{Human Evaluation}
\begin{table}[t] \small
\centering
\begin{tabular}{ccc}
\toprule
\multirow{2}{*}{\textbf{Type   of evaluation}} & \multicolumn{2}{c}{\textbf{Win Percentage}} \\ \cline{2-3} 
                                               & \textbf{SACO}     & \textbf{GPT-Speaker*}    \\ \midrule
Engagingness                                   & \bf62.9              & 37.1                    \\
Visual Relevance                               & \bf64.9              & 35.1                    \\
Personalized Relevance                         & \bf63.4              & 36.6                    \\ \bottomrule
\end{tabular}
\caption{Human Evaluation.}
\label{tab:hum}
\end{table}
To comprehensively evaluate our method, we conducted a human evaluation to compare our model and GPT-Speaker. Following \citet{Nguyen20Structural}, we considered the engagingness and relevance of captions. Engagingness evaluation considers human preference for the naturalness and appropriateness of the captions, while relevance evaluation involves visual and stylized relevance. Therefore, there are three types of evaluation. We randomly sampled 50 samples from the test set for each evaluation type above. Each sample includes an image and a style. Then, we use our model and GPT-Speaker to generate captions for these samples. We displayed the selected image-style pairs and their caption generated from our model and GPT-Speaker to 7 recruited annotators. They need to judge which captions are better quality based on the type of evaluation. As shown in Table \ref{tab:hum}, results show that the performance of our model is significantly better than GPT-Speaker, i.e., our model can generate fascinating captions.

\section{Conclusion}
In this work, we dive into the relationship between linguistic style and visual content. The first is potential visual content has varied for different styles. We propose a style-aware visual encoder that learns to capture the representation of potential visual content. 
Second, since the model is required to distinguish whether the image, style and caption are matched, we present a style-aware triplet contrast objective to improve the model's capability to discriminate triplet matching. 
In addition, we propose three novel retrieval schemes to sample positive and negative examples for contrastive learning. Results show that our method delivers new state-of-the-art performance. 

\section*{Limitations}
Although our proposed method can effectively mine latent visual content related to style, it still suffers from weaknesses in generating multiple stylized captions for the same image and style pair. Specifically, our method relies on beam search to generate diverse stylized captions for the same image and style pair and lacks the capability to control content for caption generation interactively. In further work, we will study how to generate stylized captions by interactively selecting specified regions in latent visual content.

\bibliography{ref}
\bibliographystyle{acl_natbib}

\end{document}